\documentclass[12pt,a4paper,onecolumn]{IEEEtran}

\usepackage{url}
\usepackage{graphicx}
\usepackage{color}
\usepackage[caption=false]{subfig}
\DeclareGraphicsExtensions{.jpg,.png,.mps,.tif,.pdf,.eps,.JPG} 
\usepackage{amsmath}
\usepackage{amssymb}

\DeclareMathOperator{\KIC}{KIC}
\DeclareMathOperator{\tr}{tr}
\DeclareMathOperator{\HSIC}{HSIC}
 \usepackage{url}

\begin{document}
\title{Kernel-based Information Criterion}
\author{
  {\bf Somayeh Danafar\IEEEauthorrefmark{1}\IEEEauthorrefmark{2}}, \and {\bf Kenji Fukumizu}\IEEEauthorrefmark{3}, \and {\bf Faustino Gomez}\IEEEauthorrefmark{1}\\

	\IEEEauthorrefmark{1}IDSIA/SUPSI, Manno-Lugano, Switzerland\\
	\IEEEauthorrefmark{2}Universit\`{a} della Svizzera Italiana, Lugano, Switzerland\\
	\IEEEauthorrefmark{3}The Institue of Statistical Mathematics, Tokyo, Japan\\
}

%\date{Agust 2014}
\maketitle

\begin{abstract}
  This paper introduces Kernel-based Information Criterion (KIC) for
  model selection in regression analysis. The novel kernel-based
  complexity measure in KIC efficiently computes the interdependency
  between parameters of the model using a variable-wise variance and
  yields selection of better, more robust regressors. Experimental results show
  superior performance on both simulated and real data sets compared
  to Leave-One-Out Cross-Validation (LOOCV), kernel-based
  Information Complexity (ICOMP), and maximum log of marginal
  likelihood in Gaussian Process Regression (GPR). 
\end{abstract}

\section{Introduction}
  
Model selection is an important problem in many areas including
machine learning.  If a proper model is not selected, any effort for
parameter estimation or prediction of the algorithm's outcome is
hopeless. Given a set of candidate models, the goal of model selection
is to select the model that best approximates the observed data and
captures its underlying regularities. Model selection criteria are
defined such that they strike a balance between the \emph{Goodness-of-fit
  (GoF)}, and the \emph{generalizability} or \emph{complexity} of the
models. 
\begin{align} \label{eq:MSC} \text{Model~selection~criterion}
  = \text{GoF}+ \text{Complexity}. 
\end{align}

Goodness-of-fit measures how well a model capture the regularity in
the data. Generalizability/complexity is the assessment of the performance of the model on unseen data or how accurately the model fits/predicts the future data. Models with higher complexity than necessary can suffer from overfitting and poor
generalization, while models that are too simple will 
underfit and have low GoF~\cite{Bishop}.

Cross-validation~\cite{Arlot}, bootstrapping~\cite{Kohavi}, Akaike Information Criterion (AIC)~\cite{Akaike}, and Bayesian Information Criterion (BIC)~\cite{Schwarz}, are well known examples of traditional model selection. In re-sampling methods such as cross-validation and bootstraping, the generalization error of the
model is estimated using Monte Carlo simulation. \cite{Metropolis}. In
contrast with re-sampling methods, the model selection methods like
AIC and BIC do not require validation to compute the model error, and
are computationally efficient. In these procedures an
\emph{Information Criterion} is defined such that the generalization
error is estimated by penalizing the model's error on observed data. A
large number of information criteria have been introduced with
different motivations that lead to different theoretical properties.
For instance, the tighter penalization parameter in BIC favors simpler
models, while AIC works better when the dataset has a very large
sample size.

Kernel methods are strong, computationally efficient analytical tools
that are capable of working on high dimensional data with arbitrarily
complex structure. They have been successfully applied in wide range
of applications such as classification, and regression. In kernel
methods, the data are mapped from their original space to a higher
dimensional feature space, the Reproducing Kernel Hilbert Space
(RKHS). The idea behind this mapping is to transform the nonlinear
relationships between data points in the original space into an
easy-to-compute linear learning problem in the feature space. For
example, in kernel regression the response variable is described as a
linear combination of the embedded data. 

Any algorithm that can be represented through dot products has a
kernel evaluation. This operation, called kernelization, makes it
possible to transform traditional, already proven, model selection
methods into stronger, corresponding kernel methods. The literature on
kernel methods has, however, mostly focused on kernel selection and on
tuning the kernel parameters, but only limited work being done on
kernel-based model selection~\cite{Rosipal,Kobayashi,Demyanov,Zhang}.
In this study, we investigate a kernel-based information criterion for
ridge regression models. In Kernel Ridge Regression (KRR), tuning the
ridge parameters to find the most predictive subspace with respect to
the data at hand and the unseen data is the goal of the kernel model
selection criterion.

In classical model selection methods the performance of the model
selection criterion is evaluated theoretically by providing a
consistency proof where the sample size tends to infinity and
empirically through simulated studies for finite sample sizes
\footnote{The reader is referred to Nishii \cite{Nishii} for a
  detailed discussion on the asymptotic properties of a wide range of
  information criteria for linear regression.}.
Other methods investigate a probabilistic upper bound of the generalization
error~\cite{Vapnik}. 

Proving the consistency properties of the model selection in \emph{kernel model selection} is challenging. The proof procedure of the classical methods does not work here. Some reasons for that are: the size of the model to evaluate problems such as under/overfitting~\cite{Bishop} is not apparent (for $n$ data points of dimension $p$, the kernel is $n \times n$, which is independent of $p$) and asymptotic probabilities of generalization error or estimators are hard to compute in RKHS.

Researchers have kernelized the traditional model selection criteria and shown the success of their kernel model selection empirically. Kobayashi and Komaki
\cite{Kobayashi} extracted the kernel-based Regularization Information
Criterion (KRIC) using an eigenvalue equation to set the
regularization parameters in Kernel Logistic Regression and Support
Vector Machines (SVM). Rosipal et al.~\cite{Rosipal} developed
Covariance Information Criterion (CIC) for model selection in Kernel
Principal Component Analysis, because of its outperformed results
compared to AIC and BIC in orthogonal linear regression. Demyanov et
al.~\cite{Demyanov}, provided alternative way of calculating the
likelihood function in Akaike Information Criterion
(AIC,~\cite{Akaike} and Bayesian Information Criterion
(BIC,~\cite{Schwarz}), and used it for parameter selection in SVMs
using the Gaussian kernel.

As pointed out by van Emden~\cite{vanEmden}, a desirable model is the
one with the fewest dependent variables. Thus defining a complexity term
that measures the interdependency of model parameters enables one to
select the most desirable model. In this study, we define a novel
variable-wise variance and obtain a complexity measure as the additive
combination of kernels defined on model parameters. Formalizing the
complexity term in this way effectively captures the interdependency of
each parameter of the model. We call this novel method 
\emph{Kernel-based Information Criterion (KIC)}. 

Model selection criterion in Gaussian Process Regression (GPR;~\cite{Rasmussen}), and kernel-based Information Complexity (ICOMP;~\cite{Zhang}) resemble KIC in using a covariance-based complexity measure. However, the methods differ because these complexity measures capture the interdependency between the data points rather than the
model parameters.

Although we can not establish the consistency properties of KIC theoretically,
we empirically evaluate the efficiency of KIC both on
synthetic and real datasets obtaining state-of-the-art results
compared to Leave-One-Out-Cross-Validation (LOOCV), kernel-based
ICOMP, and maximum log marginal likelihood in GPR. The paper is
organized as follows. In section~\ref{sec:krr}, we give an overview of
Kernel Ridge regression. KIC is described in detail in
section~\ref{sec:kic}. Section~\ref{sec:om} is provides a brief
explanation of the methods to which KIC is compared, and in
section~\ref{sec:exp} we evaluate the performance of KIC through sets
of experiments.

\section{Kernel Ridge Regression}
\label{sec:krr}
In regression analysis, the regression model of the form:
\begin{align}
Y &= f({\bf X},\theta)+\epsilon,
\end{align}
where $f$ can be either a linear or non-linear function.

In linear regression we have, $ Y= {\bf X}\theta+\epsilon $, where $Y$
is an observation vector (response variable) of size $n \times 1$,
${\bf X}$ is a full rank data matrix of independent variables of size
$n\times p$, and $\theta = (\theta_1,..,\theta_p)^T$, is an unknown
vector of regression parameters, where $T$ denotes the transposition.
We also assume that the error (noise) vector $\epsilon$ is an
$n$-dimensional vector  whose elements are drawn i.i.d,
$\mathcal{N}(0,\sigma^2I)$, where $I$ is an $n$-dimensional identity
matrix and $\sigma^2$ is an unknown variance.

The regression coefficients minimize the squared errors, $\|\hat{f}-f
\|^2$, between estimated function $\hat{f}$, and target function $f$.
When $p > n$, the problem is ill-posed, so that some kind of
regularization, such as Tikhanov regularization (ridge
regression) is required, and the coefficients minimize the following  optimization problem
\begin{align}
\text{argmin}~( Y-{\bf X} \theta)^T (Y-{\bf X} \theta)+\alpha \theta^T \theta,
\end{align}
where $\alpha$ is the regularization parameter. The estimated regression coefficients in ridge regression $\hat{\theta}$ are:
\begin{align}
\hat{\theta} &= ({\bf X}^T{\bf X}+\alpha I)^{-1} {\bf X}^T Y \nonumber\\ 
~~&= {\bf X}^T({\bf XX}^T+\alpha I)^{-1} Y.
\end{align}
In \emph{Kernel} Ridge Regression (KRR), the data matrix ${\bf X}$ is
non-linearly transformed in RKHS using a feature map $\phi({\bf X})$.
The estimated regression coefficients based on $\phi(\cdot)$ are:
  \begin{align}
  \label{eq:theta}
 \hat{\theta} &= \phi({\bf X})^T(K+\alpha I)^{-1} Y,
\end{align} where $K= \phi({\bf X})\phi({\bf X})^T$ is the kernel
matrix. Equation~\ref{eq:theta} does not obtain an explicit expression
for $\theta$ because of $\phi({\bf X})$ (the kernel trick enables one
to avoid explicitly defining $\phi(\cdot)$ that could be numerically
intractable if computed in RKHS, if known), thus a ridge estimator is
used (e.g. \cite{Zhang}) that excludes $\phi({\bf X})$: \begin{align}
  \theta^{\ast} &=(K+\alpha
  I)^{-1} Y. \end{align}
 Using $\theta^\ast$ in the calculation of KRR is similar to regularizing the regression function instead of the regression coefficients, where the objective function is:
\begin{align}
\hat{f} &= \text{argmin}_{f\in \mathcal{H}} ( Y-f({\bf X}))^T(y-f({\bf X}))+ \alpha \|f\|_{\mathcal{H}}^2,
\end{align}
and $\mathcal{H}$ denotes the relevant RKHS. For $f=K\theta$, and $\{(X_1,Y_1),.., (X_n,Y_n)\}$ we have:
\begin{align}
\hat{f} &= \sum_{i=1}^n \theta_i k(.,X_i),\\
\hat{\theta} &= \text{argmin}_{\hat{\theta}\in \mathbb{R}^n} \|Y-K\theta\|_2^2+ \alpha~\theta^T K \theta,
\end{align} 
where $k$ is the kernel function, and $\hat{\theta} = (K+\alpha I)^{-1} Y$.

\section{Kernel-based Information Criterion}
\label{sec:kic}
The main contribution of this study is to introduce a new Kernel-based
Information Criterion (KIC) for the model selection in kernel-based
regression. According to equation~\eqref{eq:MSC} KIC balances between
the goodness-of-fit and the complexity of the model. GoF is defined
using a log-likelihood-based function ( we maximize penalized log
likelihood) and the complexity measure is a function based on the
covariance function of the parameters of the model. In the next
subsections we elaborate on these terms.

\subsection{Complexity Measures}
The definition of van Emden \cite{vanEmden} for the complexity measure
of a random vector is based on the interactions among random variables
in the corresponding covariance matrix. A desirable model is the one
with the fewest dependent variables. This reduces the information
entropy and yields lower complexity. In this paper we focus on this
definition of the complexity measures.

Considering a $p$-variate normal distribution $f(X) = f(X_1,..,X_p)=
\mathcal{N}(\mu, \Sigma)$, the complexity of a covariance matrix,
$\Sigma$, is given by the Shannon's entropy \cite{Shannon},
\begin{align} \label{eq:complexity}
  C(\Sigma)&= \sum_{j=1}^{p}H(X_j)-H(X_1,...,X_p)\nonumber\\
  ~~&= \frac{1}{2} \sum_{j=1}^{p}
  \log(\sigma_{jj})-\frac{1}{2}\log |\Sigma|, 
\end{align} where
$H(X_j)$, $H(X_1,.., X_p)$ are the marginal and the joint entropy, and
$\sigma_{jj}$ is the $j{\text{-th}}$ diagonal element of $\Sigma$.
$C(\Sigma)=0$ if and only if the covariates are independent. The
complexity measure in equation~\eqref{eq:complexity} changes with
orthonormal transformations because it is dependent on the coordinates of
the random variable vectors $ X_1,.., X_p$ \cite{Bozdogan}. To
overcome these drawbacks, Bozodgan and Haughton \cite{Bozdogan}
introduced ICOMP information criterion with a complexity measure based
on the maximal covariance complexity, which is an upper bound on the
complexity measure in equation~\eqref{eq:complexity}: 
\begin{align}
  \label{eq:cICOMP} C(\Sigma) = \frac{1}{2}
  \log\left(\frac{\left(\frac{\tr(\Sigma)}{p}\right)^p}
    {|\Sigma|}\right) = \frac{p}{2}\log
  \left(\frac{\bar{\lambda}_a}{\bar{\lambda}_g}\right). 
\end{align}
This complexity measure is proportional to the estimated arithmetic
($\bar{\lambda}_a$) and geometric mean ($\bar{\lambda}_g$) of the
eigenvalues of the covariance matrix. Larger values of $C(\Sigma)$,
indicates higher dependency between random variables, and vice versa.
Zhang \cite{Zhang} introduced a kernel form of this complexity measure
$C(\Sigma)$, that is computed on kernel-based covariance of the ridge
estimator:
 \begin{align}
 \label{eq:SigmaICOMP}
\Sigma_{\theta^{\ast}} &= \sigma^2 (K+\alpha I)^{-2}.
\end{align}
The complexity measure in Gaussian Process Regression (GPR; \cite{Rasmussen}) is defined as $\frac{1}{2}log|\Sigma|$, a concept from the joint entropy $H(X_1,.., X_p)$ (as shown in equation~\ref{eq:complexity}). 
 
 In contrast to ICOMP and GPR, the complexity measure in KIC is defined
using the Hilbert-Schmidt (HS) norm of the covariance matrix,
$C(\Sigma)=\| \Sigma \|_{HS}^2= \tr (\Sigma^T\Sigma)$. Minimizing this
complexity measure obtains a model with more independent variables.

In the next sections, we explain in detail how to define the needed variable-wise variance in the complexity measure, and the computation of the complexity measure.\\

\subsubsection{{\bf Variable-wise Variance}}
In kernel-based model selection methods such as ICOMP, and GPR, the
complexity measure is defined on a covariance matrix that is of size
$n\times n$ for ${\bf X}$ of size $n\times p$. The idea behind this 
measure is to compute the interdependency between the model
parameters, which independent of the
number of the model parameters $p$. In the other words, the concept of
the size of the model is hidden because of the definition of a kernel.
To have a complexity measure that depends on $p$, we introduce
variable-wise variance using an additive combination of kernels for
each parameter of the model.

Let $\theta \in \mathcal{H}$ be the parameter vector of the kernel ridge regression:
\begin{align}
\theta & = Y(K+\alpha I)^{-1}{\bf k}(\cdot),
\end{align}
where $Y =(Y_1,..,Y_n)^T$ and ${\bf k}(\cdot) = (k(\cdot,X_1),...,k(\cdot,X_n))^T$, and ${\bf X}=\begin{bmatrix} X_1^1 & \cdots & X_p^1 \\ \vdots & \ddots & \vdots \\
X_1^n & \cdots & X_p^n \end{bmatrix}.$ The solution of KRR is given by $f(x)=~<\theta,k(\cdot,x)>$. The quantity $\tr[\Sigma_\theta] = \sigma^2 \tr[K(K+\alpha I)^{-2}] $ can be interpreted as the sum of variances for the component-wise parameter vectors, if the following sum of component-wise kernel is introduced:
\begin{align}
k(x,\tilde{x})& = \sum_{j=1}^p k_j(x_j,\tilde{x}_j),
\end{align}
where $x_j$ and $\tilde{x}_j$ denote the j-th component of vectors $x$
and $\tilde{x} \in \mathbb{R}^p$. With this sum kernel, the function $f \in \mathcal{H}$ can be written as:
\begin{align}
\label{eq:g}
f & = g_1(x_1)+...+g_p(x_p),
\end{align}
where $g_j$ is a function in $\mathcal{H}_j$, the RKHS defined by $k_j$.
The parameter $\theta$ in this case is given by
\begin{align}
\theta & = Y(K+\alpha I)^{-1} \Big(\sum_j {\bf k} _j(\cdot)\Big) = \sum_{j=1}^{p} \theta_j,
\end{align}
where $\theta_j = Y(K+\alpha I)^{-1} {\bf k} _j(.)$, and thus $g_j$ in equation~\ref{eq:g} is equal to $\theta_j$. Let $V_j$ be the conditional covariance of $\theta_j$ or $g_j$ given $(X^1,..,X^n)$. We have
\begin{align}
V_j &= \sigma^2 \tr[K_j (K+\alpha I)^{-2}],
\end{align}
where $K_{j,ab}= K_j(X_j^a, X_j^b)$ be the Gram matrix with $k_j$. Since $K_{ab}= \sum_{j=1}^p K_{j,ab}$, we have 
\begin{align}
\sum_{j=1}^p V_j & = \sigma^2 \tr[ K(K+\alpha I)^{-2}] = \tr[\Sigma_{\theta}].
\end{align}

Formalizing the complexity term with variable-wise variance effectively captures the interdependency of each parameter of the model (measures the significance of the contribution by the variables) explicitly.\\

\subsubsection{ {\bf Hilbert-Schmidt Independence Criterion}}

Gretton et al. \cite{Gretton} introduced a kernel-based independence measure, namely the Hilbert-Schmidt Independence criterion (HSIC), which is explained here. Suppose $X \in \mathcal{X}$, and $Y \in \mathcal{Y}$ are random vectors with feature maps $\phi: \mathcal{X} \rightarrow \mathcal{U}$, and $\psi:\mathcal{Y} \rightarrow \mathcal{V}$ , where $\mathcal{U}$, and $\mathcal{V}$ are RKHSs. The cross-covariance operator corresponding to the joint probability distribution $P_{XY}$ is a linear operator, $\Sigma_{XY}:\mathcal{U} \rightarrow \mathcal{V}$ such that:
\begin{align}
\Sigma_{XY} &:= E_{XY}[(\phi(X)-\mu_X) \otimes (\psi(Y)-\mu_Y)],
\end{align}
where $\otimes$ denotes the tensor product, $\mu_X = E_X[u(X)]= E[k(\cdot,X)]$, and $\mu_Y = E_Y[v(Y)]=E[k(\cdot,Y)]$, for $u\in \mathcal{U}, v \in \mathcal{V}$, and associated kernel function $k$. The HSIC measure for separable RKHs $\mathcal{U}$, and $\mathcal{V}$ is the squared HS-norm of the cross-covariance operator and is denoted as:
\begin{align}
\HSIC(P_{XY},\mathcal{U}, \mathcal{V})&:=\| \Sigma_{XY} \|_{HS}^2 = \tr[\Sigma_{XY}^T \Sigma_{XY}]
\end{align}
{\bf Theorem 1.}  Assume $\mathcal{X}$, and $\mathcal{Y}$ are compact, for all $u \in \mathcal{U}$, and $v \in \mathcal{V}$, $\| u \|_{\infty}\leq 1$, and $\| v \|_{\infty} \leq 1$, $\| \Sigma_{XY}\|_{HS} =0$ if and only if $X$, and $Y$ are independent (Theorem 4 in \cite{Gretton}). \\

By computing the HSIC on covariance matrix associated with model's parameters $\Sigma_{\theta}$ we can measure the independence between the parameters. Since $ \Sigma_\theta$ is a symmetric positive semi-definite matrix, $\Sigma^T\Sigma = \Sigma^2$, and the trace of the HS norm of the covariance matrix is equal to: 
\begin{align}
\label{eq:HS}
\|\Sigma_\theta\|_{HS}^2 &= \tr [\Sigma_\theta^T\Sigma_\theta] = \sum_{j=1}^p V_j^2\nonumber\\
~~&= \sigma^4 \tr[K(K+\alpha I)^{-2} K(K+\alpha I)^{-2}]
\end{align}

\subsection{Kernel-based Information Criterion}
KIC is defined as:
\begin{align}
KIC (\hat{\Sigma}_{\hat{\theta}}) &= -2~\text{Penalized log-likelihood} (\hat{\theta})+ C(\hat{\Sigma}_{\hat{\theta}}),
\end{align}
where $C(\hat{\Sigma}_{\hat{\theta}})=\|\hat{\Sigma}_{\hat{\theta}}\|_{HS}^2 /\sigma^4$ is the complexity term based on equation~\ref{eq:HS}. The normalization by $\sigma^4$ obtains a complexity measure that is robust to changes in variance (similar to ICOMP criterion). The minimum KIC defines the best model\footnote{Under the normality assumption for random errors, we can use the least-squares instead of the log-likelihood in the formula. Therefore KIC can also be written as $KIC (\hat{\Sigma}_{\hat{\theta}}) = 2~\text{ Penalized least-squares} (\hat{\theta})+ C(\hat{\Sigma}_{\hat{\theta}})$.}. 
The Penalized log-likelihood (PLL) in KRR for normally distribution data is defined by:
\begin{align}
\text{PLL}(\theta,\sigma^2)
 &= -\frac{n}{2}\log(2\pi)-\frac{n}{2}\log(\sigma^2)-\frac{(Y-K\theta)^T(Y-K\theta)}{2\sigma^2}-\alpha \left( \frac{\theta^T K\theta}{2\sigma^2}\right),
\end{align}
The unknown parameters $\hat{\theta}$, and $\widehat{\sigma^2}$ are calculated by minimizing the KIC objective function.
\begin{align}
\frac{\partial KIC}{\partial \theta}&= 0~~ \rightarrow ~~ \hat{\theta} = (K+\alpha I)^{-1}Y,\\
\frac{\partial KIC}{\partial \sigma^2}&=0~~\rightarrow ~~\widehat{\sigma^2}= \frac{(Y-K\theta)^T(Y-K\theta)+ \alpha \theta^TK\theta}{n}
\end{align}
We also investigated the effect of using $\tr[\Sigma_\theta^T \Sigma_\theta]$, and $\tr[\Sigma_\theta]$ as complexity terms. The empirical results reported in subsection~\ref{subsec:realData} on real datasets, and compared with KIC. We denote these information criteria as:

\begin{align}
\label{eq:KIC1}
\KIC\_1 & = -2~\text{PLL}+ \sigma^2 \tr[K(K+\alpha I)^ {-2}],
\end{align}
\begin{align}
\label{eq:KIC2}
\KIC\_2 & = -2~\text{PLL}+ \sigma^4 \tr[K(K+\alpha I)^{-2}K (K+\alpha I)^{-2}].
\end{align}
In both KIC\_1, and KIC\_2, similar to KIC, $\hat{\theta} = (K+\alpha I)^{-1}Y$, while because the complexity term is dependent on $\sigma^2$, $\widehat{\sigma^2}$ for KIC\_1 is:
\begin{align}
\frac{\partial KIC}{\partial  \sigma^2}&=0 \rightarrow\frac{n}{\sigma^2}-\frac{D}{\sigma^4}+ \tr[K(K+\alpha I)^{-2}]=0.
\end{align}
If we denote $\sigma^2=Z$, $\widehat{\sigma^2}$ is the solution of a quadratic optimization problem, $C(\Sigma)~Z^2+n~Z-D=0$, where $D=(Y-K\theta)^T(Y-K\theta)+ \alpha \theta^T K \theta$. In the case of KIC\_2, the $\widehat{\sigma^2}$ is the real root of the following cubic problem:
\begin{align}
2~C(\Sigma)~Z^3+ n~Z-D= 0,
\end{align}
where $C(\Sigma) =\tr[K(K+\alpha I)^{-2} K (K+\alpha I)^{-2}]$.

\section{Other Methods} 
\label{sec:om}
We compared KIC with LOOCV \cite{Wahba}, kernel-based ICOMP
\cite{Zhang}, and maximum log of marginal likelihood in GPR (abbreviated
as GPR) \cite{Rasmussen} to find the optimal ridge regressors. The
reason to compare KIC with ICOMP and GPR is that in all of these
methods the complexity measure computes the interdependency of model
parameters as a function of covariance matrix in different ways. LOOCV
is a standard and commonly used methods for model selection.

{\bf LOOCV:} Re-sampling model selection methods like cross-validation is time consuming \cite{Arlot}. For instance, the Leave-One-Out-Cross-Validation (LOOCV) has the computational cost of $l \times O(A) \times$ the number of parameter combinations ($O(A)$ is the processing time of the model selection algorithm $A$) for $n-1$ training samples. To have cross-validation methods with faster processing time, the closed form formula for the risk estimators of the algorithm under special conditions are provided. We consider the kernel-based closed form of LOOCV for linear regression introduced by \cite{Wahba}: 
\begin{align}
\text{Squared~Error}_{\text{LOOCV}} &= \frac{\|[diag(I-H)]^{-1}[I-H]Y\|_2^2}{n}
\end{align}
where $H = (K+ \alpha I)^{-1}K$ is the hat matrix.

{\bf Maximizing the log of Marginal Likelihood (GPR)} is a kernel-based regression method. For a given training set $\{x_is\}_{i=1}^n$, and $y_i= f(x_i)+\epsilon$, a multivariate Gaussian distribution is defined on any function $f$ such that, $P(f(x))= \mathcal{N}(\mu, \Sigma)$, where $\Sigma$ is a kernel.
Marginal likelihood is used as the model selection criterion in GPR, since it balances between the lack-of-fit and complexity of a model. Maximizing the log of marginal likelihood obtains the optimal parameters for model selection. The log of marginal likelihood is denoted as:
\begin{align}
\log P(y|x,\Theta) &= -\frac{1}{2}y^T\Sigma^{-1}y-\frac{1}{2}\log|\Sigma|-\frac{n}{2}\log 2\pi,
\end{align}
where $-y^T\Sigma y$ denotes the model's fit, $\log |\Sigma|$, denotes the complexity , and $\frac{n}{2}\log 2 \pi$ is a normalization constant. Without loss of generality in this paper GPR means the model selection criterion is used in GPR.

{\bf ICOMP:} The kernel-based ICOMP introduced in \cite{Zhang} is an information criterion to select the models and is defined as $ICOMP= -2~\text{PLL}+ 2~C(\Sigma)$, where $C(\Sigma)$, and $\Sigma$ elaborated in equations~\ref{eq:cICOMP}, and~\ref{eq:SigmaICOMP}.

\section{Experiments}
\label{sec:exp}
In this section we evaluate the performance of KIC on synthetic, and
real datasets, and compare with competing model selection methods.

\subsection{Artificial Dataset}

\begin{figure}[t] \centering 
\includegraphics[scale =  0.45]{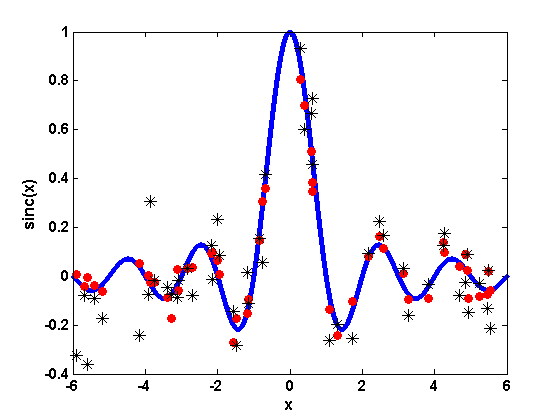} 
  \caption{The simulated sinc function is depicted with solid line, and the noisy sinc function on 50 randomly selected data points with NSR=$5\%$,
    and $40\%$ are shown respectively with red dots, and black stars.}
  \label{sinc} \end{figure}

KIC was first evaluated on the problem of approximating  $f(x)=
sinc(x)=\frac{sin(\pi x)}{\pi x}$ from a set of 100 points
sampled at regular intervals in $[-6,6]$. To evaluate robustness to
noise, normal random noise was added to the $sinc$ function at two
Noise-to-Signal (NSR) ratios: $5\%$, and $40\%$. Figure~\ref{sinc}
shows the sinc function and the perturbed datasets. The following
experiments were conducted: (1) shows how KIC balances between GoF and
complexity, (2) shows how KIC and MSE on training sets change when the
sample size and the level of noise in the data change (3) investigates the
effect of using different kernels, and (4) evaluates the consistency
of KIC in parameter selection. All experiments were run 100 times
using randomly generated datasets, and corresponding test sets of size
1000.

\begin{figure}[t] \centering \includegraphics[width= 14.3cm,
   height=4.7cm]{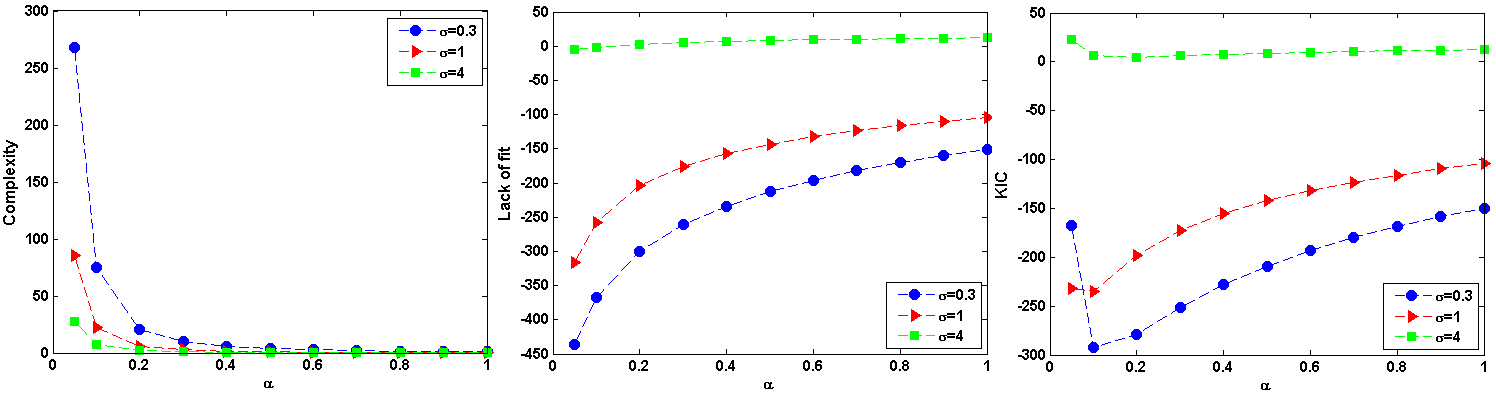} \caption{The complexity (left), goodness-of-fit (middle), and KIC values (right) of KRR models with respect to changes in
     $\alpha$ for different values of  $\sigma =\{0.3, 1, 4\}$ in
     Gaussian kernels. KIC balances between the complexity and the
     lack of fit.} \label{co_la_kic} \end{figure}

 {\bf Experiment 1.} The effect of $\alpha$ on complexity, lack-of-fit and KIC values was measured by setting $\alpha = \{ 0.05, 0.1, 0.2, 0.3, 0.4, 0.5, 0.6, 0.7, 0.8, 0.9, 1.0\}$, with KRR models being generated using a Gaussian kernel with
 different standard deviations, $\sigma= \{0.3 , 1, 4\}$, computed
 over the 100 data points. The results are shown in
 Figure~\ref{co_la_kic}. The model generated with $\sigma = 0.3$
 overfits, because it is overly complex, while $\sigma = 4$ gives a
 simpler model that underfits. As the ridge parameter $\alpha$ increases, the model complexity decreases while the goodness-of-fit is adversely affected.

KIC balances between these two terms, which
 yields a criterion to select a model that has good generalization, 
as well as goodness of fit to the data.

\begin{figure}[t]
\centering
 \subfloat[\scriptsize{(KIC, $50$)}]{\includegraphics[scale=0.45]{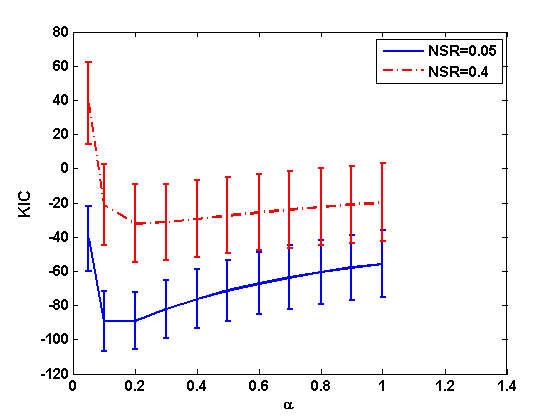}}
\subfloat[\scriptsize{(MSE, $50$)}]{\includegraphics[scale=0.45]{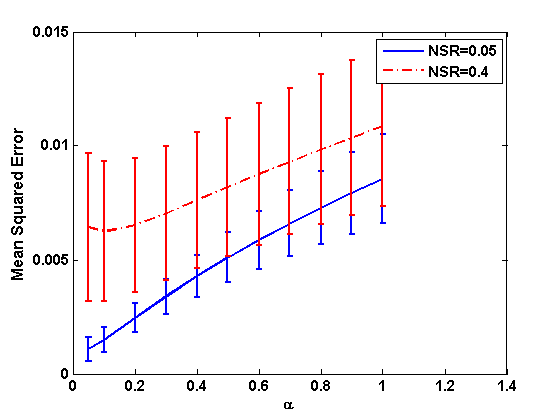}}
\hfill
 \subfloat[\scriptsize{(KIC, $100$)}]{\includegraphics[scale=0.45]{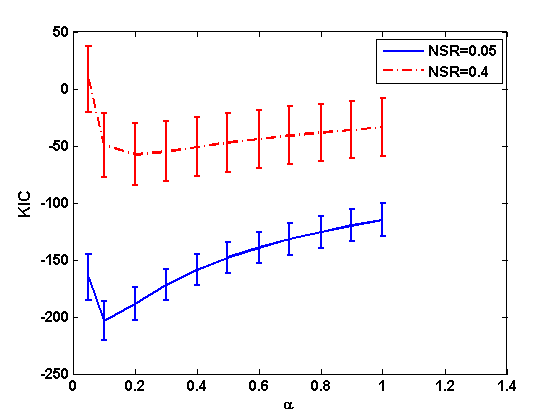}}
  \subfloat[\scriptsize{(MSE, $100$)}]{\includegraphics[scale=0.45]{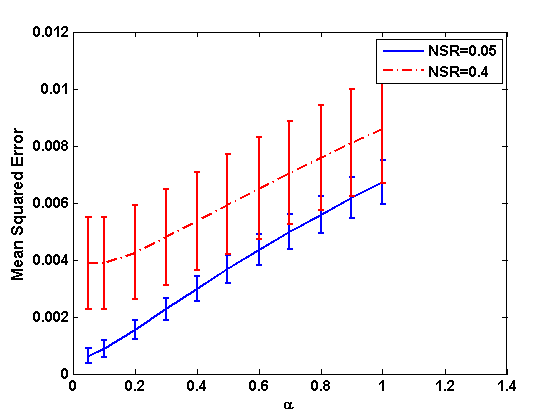}}
 \caption{(a), and (b) represent changes of KIC, and MSE values with respect to $\alpha$ for $50$ data points respectively, and (c) and (d) are corresponding diagrams for the data with $100$ sample size. The solid lines represent NSR=$0.05$, and dashed lines NSR=$0.4$.}
 \label{KIC-MSE}
 \end{figure}

 {\bf Experiment 2.} The influence of training sample size was
 investigated by comparing sample sizes, $n$, of 50, and 100, for a total of
 four sets of experiments:  ($n, \text{NSR}$): ($50, 5\%$), ($50, 40\%$), ($100, 5\%$), ($100, 40
 \%$). The Gaussian kernel was used with $\sigma = 1$. The KIC value
 and Mean Squared Error (MSE, $\|\hat{f}-f\|^2/n$), for different
 $\alpha = \{0.05, 0.1, 0.2, 0.3, 0.4, 0.5, 0.6,$ $ 0.7, 0.8, 0.9,
 1.0\}$ is shown in Figure~\ref{KIC-MSE}. The data with NSR=$40\%$ has
 larger MSE values, and larger error bars, and consequently larger KIC
 values compared to data with NSR=$5\%$. In both cases, KIC and MSE
 change with similar profiles with respect to $\alpha$. The noise and the sample size have no effect on KIC for selecting the best model (parameter $\alpha$).

\begin{figure*}[t]
  \centering
   \subfloat[\scriptsize {(Gaussian, $50, 0.05$)}]{\includegraphics[scale=0.45]{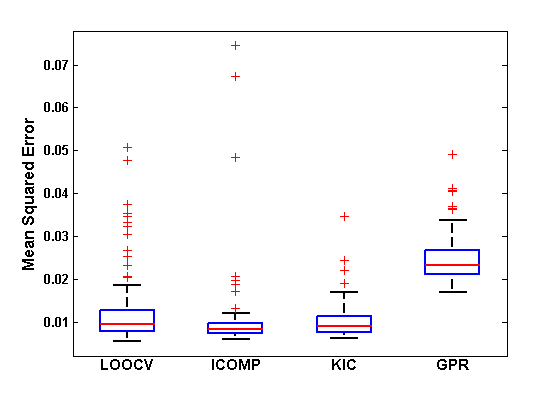}}
  \subfloat[\scriptsize{(Gaussian, $50, 0.4$)}]{\includegraphics[scale=0.45]{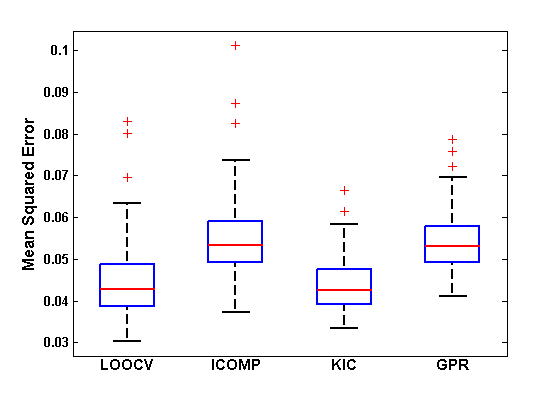}}
\hfill
\subfloat[\scriptsize{(Gaussian, $100, 0.05$)}]{\includegraphics[scale=0.45]{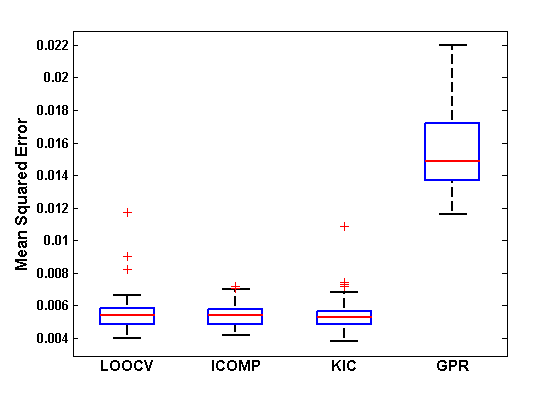}}
   \subfloat[\scriptsize{(Gaussian, $100, 0.4$)}]{\includegraphics[scale=0.45]{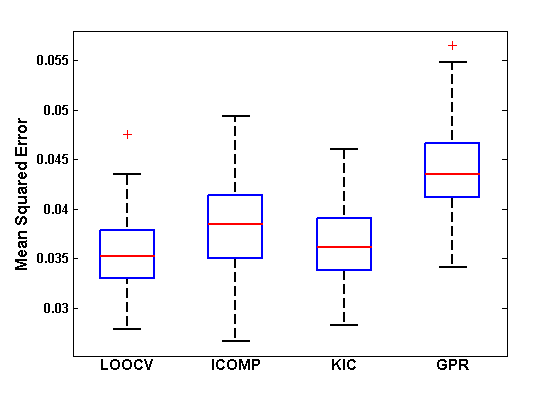}}
  \caption{The graphs depict the Mean Squared Error values as results of using Gaussian kernel with $\sigma=1$, in ICOMP, KIC, and GPR, compared with MSE of LOOCV. The results on simulated data with $(n, \text{NSR}): (50, 0.05), (50, 0.4), (100, 0.05)$, and $(100,0.4)$ are shown in (a), (b), (c), and (d) respectively. KIC gives the best results.}
 \label{Gaussian kernel}
 \end{figure*}

\begin{figure}[t]
 \centering
 \subfloat[\scriptsize {(Cauchy, $50, 0.05$)}]{\includegraphics[scale=0.45]{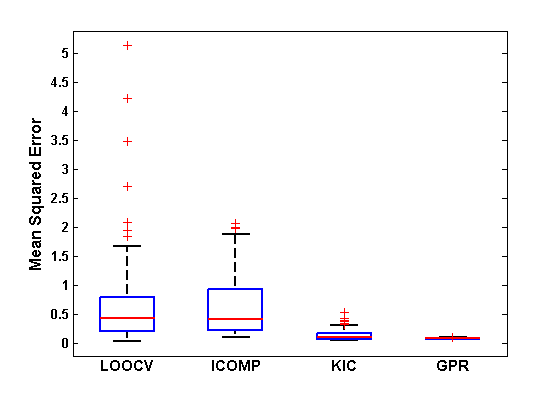}}
\subfloat[\scriptsize{(Cauchy, $50, 0.4$)}]{\includegraphics[scale=0.45]{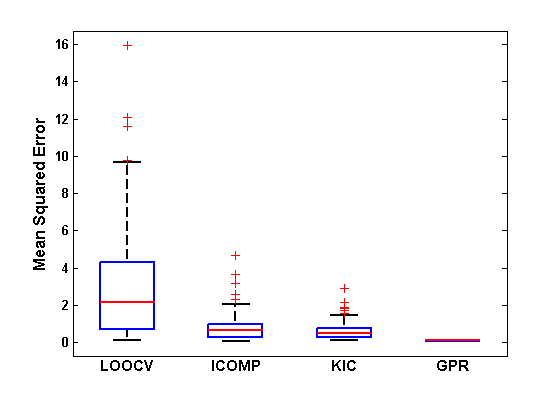}}
\hfill
\subfloat[\scriptsize{(Cauchy, $100, 0.05$)}]{\includegraphics[scale=0.45]{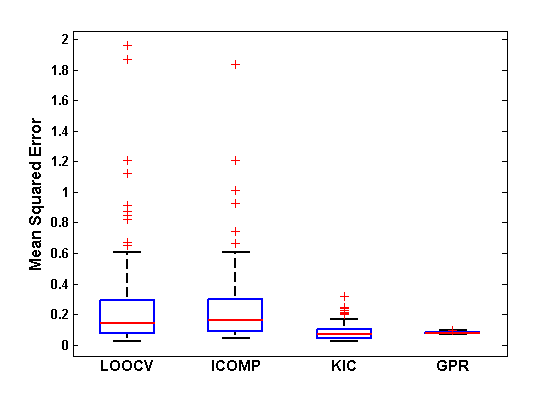}}
  \subfloat[\scriptsize{(Cauchy, $100, 0.4$)}]{\includegraphics[scale=0.45]{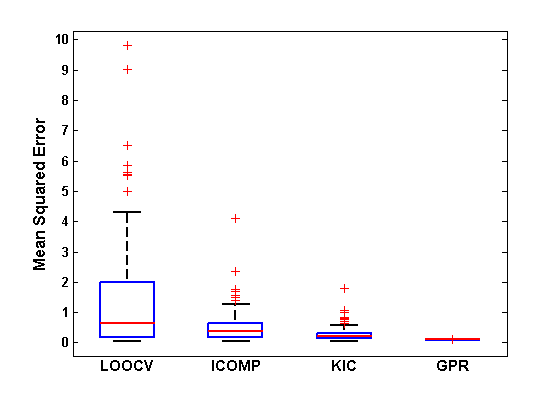}}
 \caption{The graphs show the MSE values resulting from using a Cauchy kernel with $\eta=2$, in ICOMP, KIC, and GPR, compared with the MSE of LOOCV. The results on simulated data with $(n, \text{NSR}): (50, 0.05), (50, 0.4), (100, 0.05)$, and $(100,0.4)$ are shown in (a), (b), (C), and (d) respectively.}
\label{Cauchy kernel}
\end{figure}

 {\bf Experiment 3.}  The effect of using a Gaussian
 kernel, $k(x,y) = exp (-\frac{\|x-y\|^2}{\sigma^2})$, versus the
 Cauchy kernel,
$k(x,y)=1/(1+(\frac{\|x^\eta-y^\eta\|^2}{\eta}))$,
 was investigated, where
 $\sigma=1$, and $\eta=2$ in the computation of the kernel-based model
 selection criteria ICOMP, KIC, GPR, and LOOCV. The results are
 reported in Figures~\ref{Gaussian kernel} and ~\ref{Cauchy kernel}.
 The graphs show box plots with markers at $5\%, 25\%, 50\%$, and
 $95\%$ of the empirical distributions of MSE values. As expected, the
 MSE of all methods is larger when NSR is high, $0.4$, and smaller for
 the larger of the two training sets (100 samples). LOOCV, ICOMP, and
 KIC performed comparably, and better than GPR using a Gaussian kernel
 for data with NSR $= 0.05$. In the other cases, the best results
 (smallest MSE) was achieved by KIC. All methods have smaller MSE
 values using the Gaussian kernel versus the Cauchy kernel. GPR with the
 Cauchy kernel obtains results comparable with KIC, but with a standard
 deviation close to zero.

\begin{figure*}[t]
\centering
\subfloat[\scriptsize {(LOOCV, $50, 0.05$)}]{\includegraphics[scale=0.45]{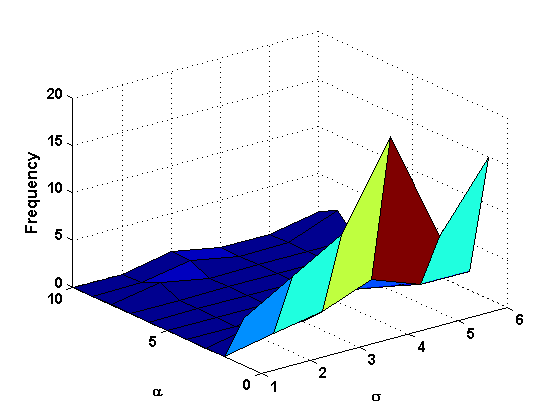}}
\subfloat[\scriptsize{(LOOCV, $50, 0.4$)}]{\includegraphics[scale =0.45]{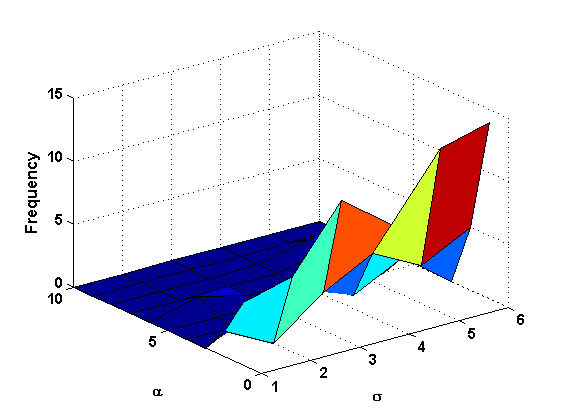}}
\hfill
\subfloat[\scriptsize{(LOOCV, $100, 0.05$)}]{\includegraphics[scale =0.45]{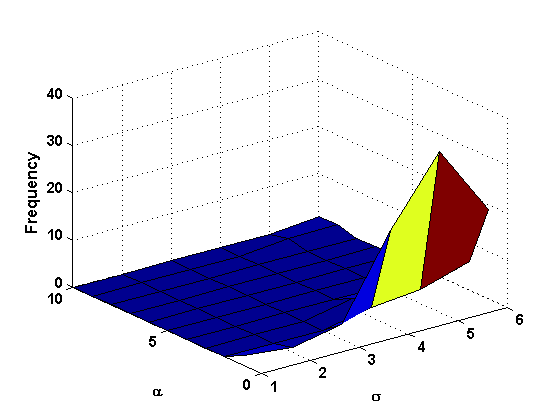}}
\subfloat[\scriptsize{(LOOCV, $100, 0.4$)}]{\includegraphics[scale=0.45]{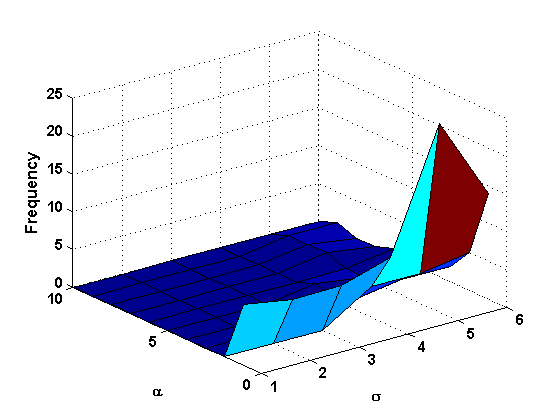}}
 \caption{The frequency of selecting parameters $\alpha$, and $\sigma$ by LOOCV method via running $100$ trials on artificial benchmarks with $(n,\text{NSR}) = (50, 0.05), (50, 0.4), (100, 0.05), (100,0.4)$ are shown in (a), (b), (c), and (d) respectively.}
\label{LOOCV}
\end{figure*}

\begin{figure*}[t]
\centering
\subfloat[\scriptsize {(KIC, $50, 0.05$)}]{\includegraphics[scale =0.45]{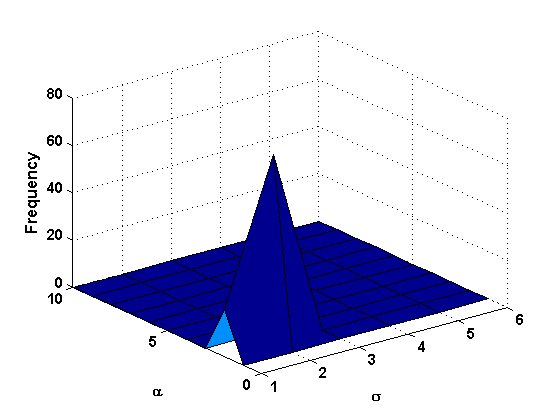}}
\subfloat[\scriptsize{(KIC, $50, 0.4$)}]{\includegraphics[scale =0.45]{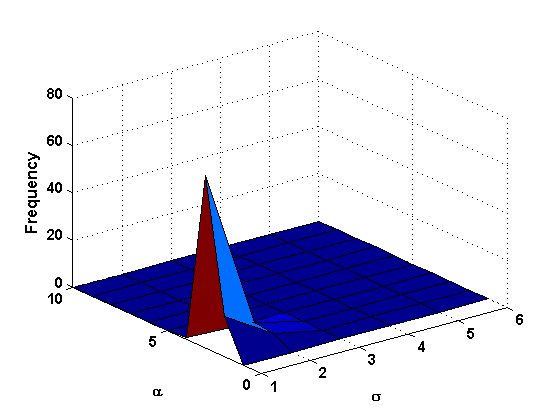}}
\hfill
\subfloat[\scriptsize{(KIC, $100, 0.05$)}]{\includegraphics[scale =0.45]{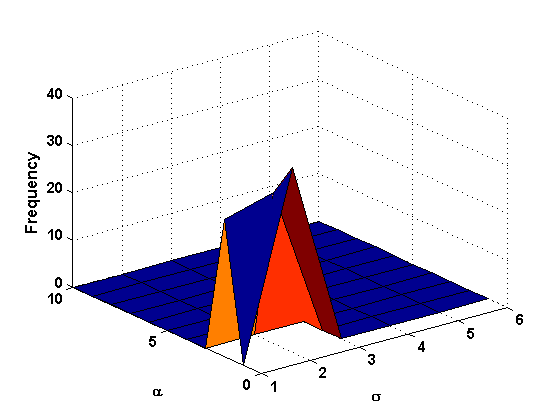}}
\subfloat[\scriptsize{(KIC, $100, 0.4$)}]{\includegraphics[scale =0.45]{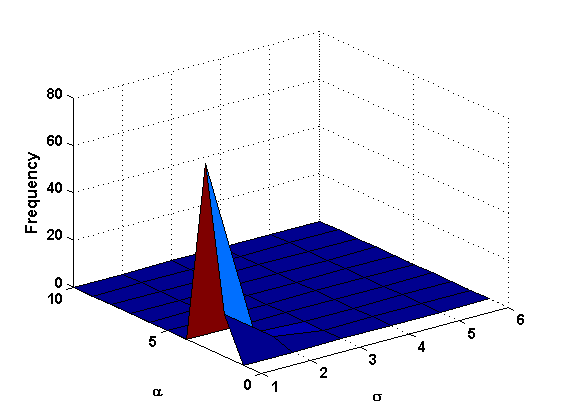}}
 \caption{The frequency of selecting parameters $\alpha$, and $\sigma$  by KIC via running $100$ trials on artificial benchmarks with $(n,\text{NSR}) = (50, 0.05), (50, 0.4), (100, 0.05), (100,0.4)$ are shown in (a), (b), (c), and (d) respectively.}
\label{KIC_frequency}
\end{figure*}

{\bf Experiment 4.} We assessed the consistency of selecting/tuning the parameters of the models in comparison with LOOCV. We considered four experiment of sample size, $n=(50,100)$, and NSR $= (0.05, 0.4)$. The parameters to tune or select are $\alpha =\{ 0.1, 0.2, 0.3, 0.4, 0.5, 0.6, 0.7, 0.8,$ $ 0.9, 1\}$, and $\sigma = \{ 0.4, 0.6, 0.8, 1, 1.2, 1.5 \}$ for the Gaussian kernel. The frequency of selecting the parameters are shown in Figure~\ref{LOOCV} for LOOCV, and in Figure~\ref{KIC_frequency} for KIC. The more concentrated frequency shows the more consistent selecting criterion. The diagrams show that KIC is more consistent in selecting the parameters rather than LOOCV. LOOCV is also sensitive to sample size. It provides a more consistent result for benchmarks with $100$ samples.\\ 

\subsection{Abalon, Kin, and Puma Datasets}
\label{subsec:realData}

We used three benchmarks selected from the Delve datasets
(\url{www.cs.toronto.edu/~delve/data}): (1) Abalone dataset (4177
instances, 7 dimensions), (2) Kin-family of datasets (4 datasets; 8192
instances, 8 dimensions), and (3) Puma-family of datasets (4 datasets;
8192 instances, 8 dimensions).

For the Abalone  dataset, the task is to estimate the age of
abalones. We used  normalized attributes in range [0,1]. The experiment
is repeated 100 times to obtain the confidence interval. In each trial
100 samples were selected randomly as the training set and the
remaining 4077 samples as the test set. The Kin-family and Puma-family
datasets are realistic simulations of a robot arm taking into
consideration combinations of attributes such as whether the arm
movement is nonlinear (n) or fairly linear (f), and whether the level
of noise (unpredictability) in the data is: medium (m), or high (h). The
Kin-family includes: kin-8fm, kin-8fh, kin-8nm, kin-8nh datasets, and
the Puma-family contains: puma-8fm, puma-8fh, puma-8nm, and puma-8nh
datasets.

In the Kin-family of datasets, having the angular positions of an 8-link
robot arm, the distance of the end effector of the robot arm from a
starting position is predicted. The angular position of a link of the
robot arm is predicted given the angular positions, angular
velocities, and the torques of the links.

\begin{figure}[t]
\centering
\includegraphics[scale = 0.5]{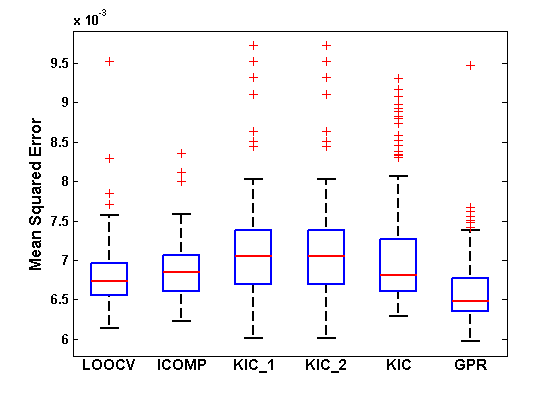}
\caption{The results of LOOCV, ICOMP, KIC\_1, KIC\_2, KIC,
  and GPR on Abalone dataset. Comparable results achieved by LOOCV,
  and KIC, with smaller MSE value rather than ICOMP, and the best
  results by GPR.}
\label{abalone}
\end{figure}

\begin{figure*}[t]
  \centering
  \subfloat[\scriptsize {kin-8fm}]{\includegraphics[scale =0.5]{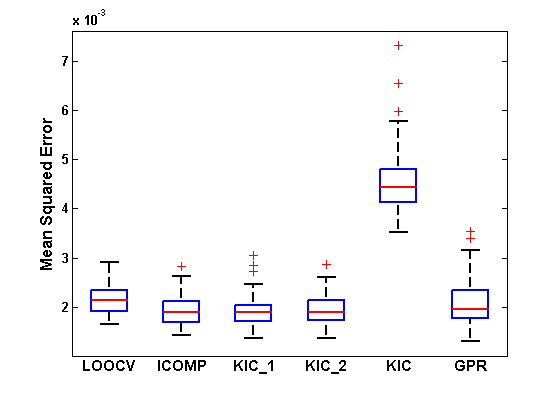}}
  \subfloat[\scriptsize{kin-8fh}]{\includegraphics[scale =0.5]{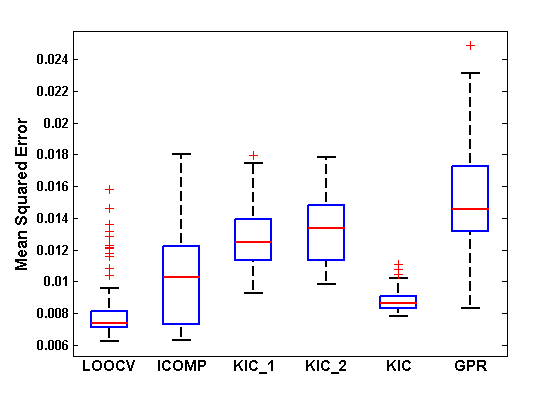}}
  \hfill
 \subfloat[\scriptsize{kin-8nm}]{\includegraphics[scale =0.5]{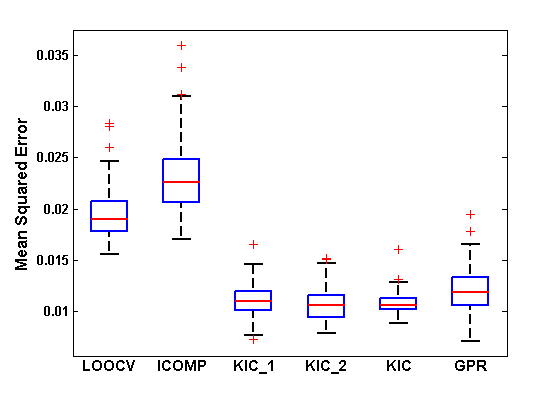}}
   \subfloat[\scriptsize{kin-8nh}]{\includegraphics[scale =0.5]{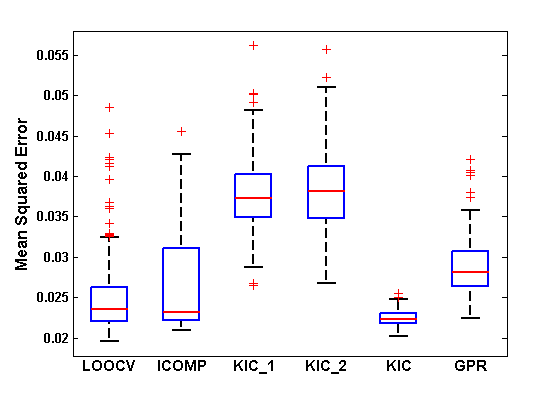}}
   \caption{The results of LOOCV, ICOMP, KIC\_1, KIC\_2, KIC, and
     GPR on kin-family of datasets are shown using box
     plots. (a),(b),(c), and (d) are the results on kin-8fm, kin-8fh,
     kin-8nm, kin-8nh, respectively.}
  \label{kin-family}
 \end{figure*}

\begin{figure*}[t]
  \centering
   \subfloat[\scriptsize {puma-8fm}]{\includegraphics[scale =0.5]{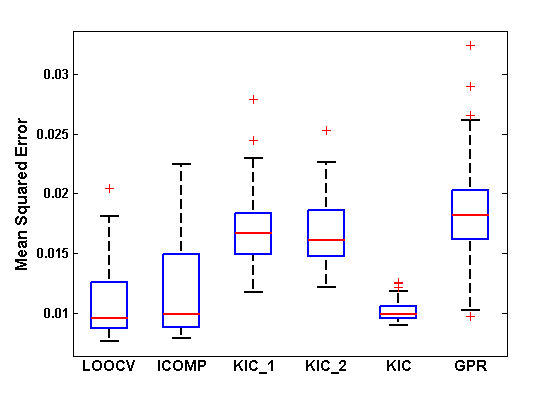}}
  \subfloat[\scriptsize{puma-8fh}]{\includegraphics[scale =0.5]{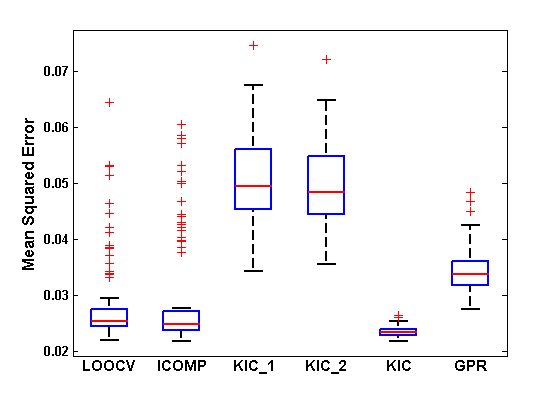}}
  \hfill
 \subfloat[\scriptsize{puma-8nm}]{\includegraphics[scale =0.5]{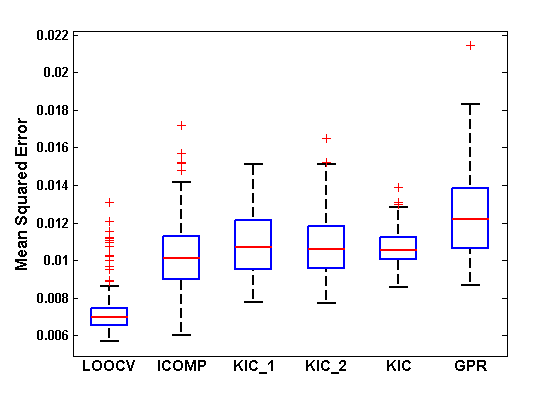}}
  \subfloat[\scriptsize{puma-8nh}]{\includegraphics[scale =0.5]{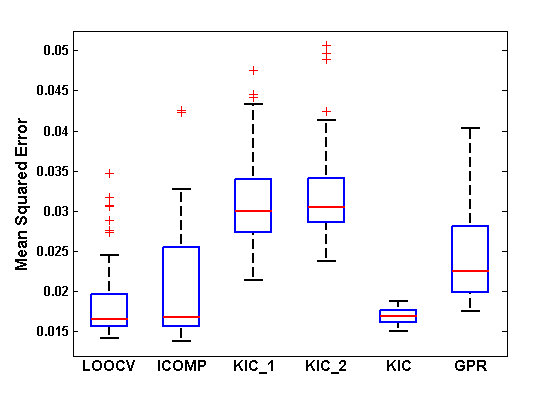}}
  \caption{The results of LOOCV, ICOMP, KIC\_1, KIC\_2, KIC, and GPR on puma-family of datasets are shown using box plots. (a),(b),(c), and (d) are the results on puma-8fm, puma-8fh, puma-8nm, puma-8nh, respectively.}
  \label{puma-family}
\end{figure*}

We compared KIC\_1 (\ref{eq:KIC1}), KIC\_2 (\ref{eq:KIC2}), and KIC
with LOOCV, ICOMP, and GPR on the three datasets. The results are
shown as box-plots in Figures~\ref{abalone}, \ref{kin-family},
and~\ref{puma-family} for Abalone, Kin-family, and Puma-family
datasets, respectively. The best results across all three datasets
were achieved using KIC, and the second best results were for
LOOCV.

For the Abalone dataset, comparable results were achieved for KIC and LOOCV,
that are better than ICOMP, and the smallest MSE value obtained by
sGPR. KIC\_1, and KIC\_2 had similar MSE values, which are larger
than for the other methods.

For the Kin-family datasets, except for kin-8fm, KIC gets better results
than GPR, ICOMP, and LOOCV. KIC\_1, and KIC\_2 obtain better results
than GPR, and LOOCV for kin-8fm, and kin-8nm, which are datasets with
medium level of noise, but larger MSE value for datasets with high
noise (kin-8fh, and kin-8nh).

For the Puma-family datasets, KIC got the best results on all datasets
except for on puma-8nm, where the smallest MSE was achieved by LOOCV. The
result of KIC is comparable to ICOMP and better than GPR for puma-8nm
dataset.  For puma-8fm, puma-8fh, and puma-8nh, although the median of
MSE for LOOCV and GPR are comparable to KIC, KIC has a more
significant MSE (smaller interquartile in the box bots). The median
MSE value for KIC\_1, and KIC\_2 are closer to the median MSE values of
the other methods on puma-8fm, and puma-8nm, where the noise level 
is moderate compared to  puma-8fh, and puma-8nh, where the
noise level is high. The sensitivity of KIC\_1, and
KIC\_2 to noise is due to the existence of variance
in their formula. KIC\_2 has a larger interquartile of MSE than KIC\_1
in datasets with high noise, which highlights the effect of $\sigma^4$
in its formula (equation ~\ref{eq:KIC2}) rather than $\sigma^2$ in
equation~\eqref{eq:KIC1}.


\section{Conclusion}
We introduced a novel kernel-based information criterion (KIC) for
model selection in regression analysis. The complexity measure in KIC
is defined on a variable-wise variance which explicitly computes the
interdependency of each parameter involved in the model; whereas in
methods such as kernel-based ICOMP and GPR, this interdependency is
defined on a covariance matrix, which obscures the true contribution of
the model parameters. We provided empirical evidence showing how KIC
outperforms LOOCV (with kernel-based closed form formula of the
estimator), Kernel-based ICOMP, and GPR, on both artificial data and
real benchmark datasets: Abalon, Kin family, and Puma family. In these
experiments, KIC efficiently balances the goodness of fit and
complexity of the model, is robust to noise (although for higher noise
we have larger confidence interval as expected) and sample size, is
consistent in tuning/selecting the ridge and kernel parameters, and
has significantly smaller or comparable Mean Squared values with
respect to competing methods, while yielding stronger regressors. The
effect of using different kernels was also investigated since the
definition of a proper kernel plays an important role in kernel
methods. KIC had superior performance using different kernels and for
the proper one obtains smaller MSE.

\section{Acknowledgment}
This work was funded by FNSNF grants (P1TIP2\_148352, PBTIP2\_140015). We want to thank Arthur Gretton, and Zolt\'{a}n Szab\'{o} for the fruitful discussions.

%\bibliographystyle{plain}
%\bibliography{biblio}
%\bibliography{bibliography}

\end{document}